\documentclass[10pt, a4paper]{article}
\usepackage{lrec2022} 
\usepackage{multibib}
\newcites{languageresource}{Language Resources}
\usepackage{graphicx}
\usepackage{tabularx}
\usepackage{soul}
\usepackage{enumitem}
\usepackage{multirow}
\usepackage{booktabs}
\usepackage[utf8]{inputenc}
\usepackage{subfig}
\usepackage{arydshln}
\usepackage{textcomp}
\usepackage[usenames, dvipsnames]{xcolor}
\usepackage{array}
\usepackage{epstopdf}
\usepackage[utf8]{inputenc}
\usepackage{hyperref}
\usepackage{xstring}
\usepackage{color}

\setlist{nolistsep}
\usepackage{titlesec}
\titleformat{\section}{\normalfont\large\bfseries\center}{\thesection.}{1em}{}
\titleformat{\subsection}{\normalfont\SmallTitleFont\bfseries\raggedright}{\thesubsection.}{1em}{}
\titleformat{\subsubsection}{\normalfont\normalsize\bfseries\raggedright}{\thesubsubsection.}{1em}{}
\renewcommand\thesection{\arabic{section}}
\renewcommand\thesubsection{\thesection.\arabic{subsection}}
\renewcommand\thesubsubsection{\thesubsection.\arabic{subsubsection}}

\title{Investigating User Radicalization: \\
A Novel Dataset for Identifying Fine-Grained Temporal Shifts in Opinion}

\name{Flora Sakketou$^\dagger$, Allison Lahnala$^\dagger$, Liane Vogel$^\ddagger$, Lucie Flek$^\dagger$} 

\address{$^\dagger$Conversational AI and Social Analytics (CAISA) Lab\\
Department of Mathematics and Computer Science, University of Marburg\\ 
$^\ddagger$Data Management Lab, Department of Computer Science,
        Technical University of Darmstadt\\
         \{flora.sakketou, allison.lahnala, lucie.flek\}@uni-marburg.de, liane.vogel@cs.tu-darmstadt.de \\}

\abstract{
There is an increasing need for the ability to model fine-grained opinion shifts of social media users, as concerns about the potential polarizing social effects increase. However, the lack of publicly available datasets that are suitable for the task presents a major challenge. In this paper, we introduce an innovative annotated dataset for modeling subtle opinion fluctuations and detecting fine-grained stances. The dataset includes a sufficient amount of stance polarity and intensity labels per user over time and within entire conversational threads, thus making subtle opinion fluctuations detectable both in long term and in short term. All posts are annotated by non-experts and a significant portion of the data is also annotated by experts. We provide a strategy for recruiting suitable non-experts. Our analysis of the inter-annotator agreements shows that the resulting annotations obtained from the majority vote of the non-experts are of comparable quality to the annotations of the experts. We provide analyses of the stance evolution in short term and long term levels, a comparison of language usage between users with vacillating and resolute attitudes, and fine-grained stance detection baselines.
 \\ \newline  }

\begin{document}

\maketitleabstract

\section{Introduction}

Social media networks play a major role in platforming the expression of opinions on innumerable topics. With concerns growing about how they may facilitate ``echo-chambers" that could could amplify polarization and radicalization \cite{delvicario2016modeling,fd84e41afabf4ffdbe90fb4f3a398eaf,10.1287/orsc.1120.0767,10.1371/journal.pone.0074516}, temporal data about user opinions is necessary in order to study the dynamics of these spaces and their potential to radicalize individuals~\cite{10.1093/ijpor/edu032}. 

While there is abundant research in sociopolitical sciences regarding opinion formation and expression, there is a significant lack of Natural Language Processing (NLP) data resources that enable the study of temporal opinion dynamics, thus also, a lack of efficient, data-driven models that investigate the relationships between social media behaviors and opinion formation and fluctuations~\cite{doi:10.1177/1532673X19832543,RePEc:spr:dyngam:v:1:y:2011:i:1:p:3-49}. Meanwhile, the increasing number of people using social media platforms makes the need for models that analyze and forecast public opinion all the more vital.

A relevant body of work concerns \textit{automatic stance detection}~\cite{mohammad-etal-2016-semeval,Walker_thatsyour,somasundaran-wiebe-2010-recognizing,thomas-etal-2006-get}, the task of determining the polarity of a stance toward a given subject, for which many datasets have been developed. These datasets however do not satisfy the needs for studying opinion dynamics from text. One need is temporal data about user stances to be able to study opinion changes or fluctuations, yet many datasets do not link stance expressions to unique authors, or otherwise the opinions are often treated as a stable entity per author. Additionally, most datasets do not contain information about the intensity of the stance. This is especially important for studying opinion dynamics, as opinion fluctuations are usually subtle--people do not tend to change the polarity of their stance but rather the intensity ~\cite{Nyhan2010,delvicario2016modeling,Jager05uniformitybipolarization,Flache2019}.

\begin{table}[t]
    \centering
    \begin{tabular}{p{\linewidth}}
         \textit{``Fellow environmentalists, join me in embracing nuclear power.'' }\\
         \textbf{Stance: {\color{ForestGreen}Strongly favors} nuclear energy} \\
        \midrule
         \textit{``The people that are anti-gun don't generally come to that position due to a strong grasp of logic and facts.''} \\
         \textbf{Stance: {\color{ForestGreen}Strongly favors} guns}\\
        \midrule
         \textit{``Vegan claims children can be healthy as vegans but doesn't prove claims and or acknowledge that healthy people that eat fast food and exercise.''} \\
         \textbf{Stance:  {\color{YellowOrange} Weakly against} veganism}\\
        \cdashline{1-1}
    \end{tabular}
    \caption{Stance Examples}
    \label{tab:stance_examples}
\end{table}

In this paper, we introduce an innovative annotated dataset, called SPINOS\footnote{https://github.com/caisa-lab/SPINOS-dataset} (\textbf{S}ubtle \textbf{P}olarity and \textbf{I}ntensity \textbf{O}pinion \textbf{S}hifts), of 3.5k Reddit entries for  modeling opinion dynamics and detecting fine-grained stances. The dataset contains more than 11k manual, quality-controlled annotations from both experts and non-experts. The dataset contains opinions on various sociopolitical topics such as abortion or gun control. It includes  a sufficient amount of stances per user on complex sociopolitical topics (1) over time, and (2) within entire conversational threads, and additionally (3) beyond stance polarity, it contains stance intensity labels, thus making subtle opinion fluctuations detectable. By introducing this dataset, we are simultaneously addressing the need of user-based diachronic stance intensity labels and bridging the gap between the theoretical sociopolitical approaches and NLP methodologies.

Our first innovation is that we provide a sufficient amount of annotated user history which enables the investigation of opinion shifts both in the long-term but also within a conversation. More specifically, we approach this in a user-centric manner, meaning for each author, we collect annotations for multiple posts where they express their stance on a specific topic. This approach enables the dynamical exploration of the fluctuations in opinion for each individual user in a macroscopic level over a long period of time. 

Following this motivation on a microscopic level, our second innovation is the annotation of all stances on every post within a conversational thread. This enables us to explore the evolution of opinion within discussions i.e. whether the authors tend to  polarize, reach a consensus or  diverge from the original trajectory of the subject to avoid conflict. Additionally, it facilitates investigations into the consistency of user stances that can provide insight into their temporal behavior. 

The third innovation is stance intensity annotations. With the combination of our other innovations, not only are historical and conversational level opinion dynamics open for investigation, but so are subtle fluctuations which are more likely to occur. 
Identifying stance intensity is a particularly demanding task, even for humans, because it requires a high level of language understanding and detecting intrinsic details in language usage, such annotation also depends on inherent biases and cultural background. In addition, broader knowledge about the issues is often necessary.

In this paper, we hope to provide a guide for future researchers who want to extend the dataset or expand to other platforms and issues, based on insights we gained from multiple phases of the annotation process. Through these phases we provide a strategy for recruiting reliable non-expert annotators by evaluating their individual performance versus expert annotators. We show that this strategy results in high quality annotations from non-experts striking a balance between cost and quality. In addition, we show via an analysis of the annotations that some of the topics are associated with much better agreement on polarity and extremity than others, and we attempt to offer insight into why. 

Through a case study, we investigate the opinion fluctuations of individual users both within their whole posting history but also within a conversation. We analyze users with fluctuating opinions and show whether there is a permanent shift in opinion polarity, when such a phenomenon occurs. Furthermore, we provide a comparative lexicon-based linguistic analysis between users with vacillating and resolute attitudes, showing which word categories are associated with each user type. In addition to analyzing the temporal stances, we also provide a classical fine-grained stance detection benchmark on various setups.

\section{Related Work}

Most available datasets for stance detection only indicate the polarity of the stance against a target (in favor, against, neutral) \cite{mohammad-etal-2016-dataset,stab-etal-2018-cross}, however recent work has aimed to annotate the stance intensity as well. \newcite{sirrianni-etal-2020-agreement} introduced a dataset for agreement prediction where the aim is to predict a post’s stance towards its parent post in a 5-point scale and \newcite{7078580} present a corpus of spontaneous, conversational speech with stance annotations on a 7-point scale. Both datasets contain statement-response pairs from conversational settings and are more suitable for detecting whether two topic-independent statements are in agreement, rather than detecting a stance from one statement on a sociopolitical issue. Moreover, they are not suitable for longitudinal studies and user-centric approaches as they do not include the users' historical posts.

In contrast, datasets by \newcite{walker-etal-2012-corpus} and \newcite{durmus-cardie-2018-exploring} contain stances on specific issues, and capture stances for particular users. While \newcite{walker-etal-2012-corpus}'s corpus contains explicit user stance labels on posts within a debate, all posts by a specific user within a debate have the same stance. Some of these users have indicated their stance on a number of debates across various topics, thus some historical user data can be extracted in such cases, but multiple stances labels on a particular issue are not guaranteed. \newcite{durmus-cardie-2018-exploring}'s dataset contains user stances before and after a debate on various topics, therefore micro-fluctuations in opinion can be observed. However, there is no way to clarify if these shifts have a temporary or long-term effect, and other research suggests that it is more likely the users would return to their original stance~\cite{10.5555/3237383.3237857,dowling_henderson_miller_2019}. Neither of these datasets contain stance intensity annotations which are essential for modeling the users' behavior since it is less likely to observe a complete change of the opinion polarity \cite{Nyhan2010,pub.1052753222,EDWARDS1996,Redlawsk_hotcognition,https://doi.org/10.1111/j.1540-5907.2006.00214.x}, and more likely to observe subtle changes in opinion intensity \cite{fd84e41afabf4ffdbe90fb4f3a398eaf}. 

In our dataset, we attempt to address these gaps with the innovative stance intensity labels on specific issues for an individual user over time. Moreover, our data is set in discussions with complex structures, as opposed to settings where argument structure is ingrained in the online platform. This is valuable because while such unstructured data adds additional challenges for annotation that require intense effort, the resulting dataset enables discourse analysis over more complex stance expressions in realistic and prevalent conversational settings taking place on social media platforms.

\section{Annotation}

We annotated the stance polarity and intensity of a users' post against a given topic, where the topic of each post is determined based on the subreddit it was posted on. In this section, we define the stance polarities and stance intensities, describe additional annotations provided in the dataset, and present the annotation procedure that was followed.

\subsection{Stance Polarity}

\paragraph{In favor/Against.} The \textit{in favor (or against)} label is assigned when the statement is:
explicitly in favor of (or against) the given topic, in support of someone/something that is in favor of (or against) the given topic, opposed to someone/something that is in favor of (or against) the given topic, mirroring someone else’s positive (or negative) opinion or when it can be implicitly derived that the statement is in favor of (or against) a given topic based on the information it contains (e.g. title of a news article, a scientific study, etc.), even though it is not in favor or against someone/something.

\paragraph{Stance not inferrable.} We do not provide a separate label for a neutral stance since, after the first deployment of the study, we observed that the neutral posts were annotated with the \textit{stance not inferrable} label and vice versa. The \textit{stance not inferrable} label is assigned when the post is irrelevant to the topic or not understandable. It is also assigned when the post is related to the topic and a stance cannot be inferred, for instance if the author asked a question (e.g. if they asked for information, opinion, orientation, confirmation) without revealing/implying the stance on the corresponding issue. We also provide labels for each subcategory of the no stance case. More specifically: (a) the post is (somewhat) related to the topic, but a stance cannot be inferred, (b) the author asks a question without revealing a stance, (c) the post is irrelevant to the target topic, (d) the post is not understandable.

\subsection{Stance Intensity}
When an annotator chooses the polarity of the stance, they also select the intensity as either strong or weak.

\paragraph{Strong.} The statements are annotated with strong intensity if they express strong emotions or passion about the corresponding issue, or the author of the statement seems to be adamant in their opinion. They are also labeled strong if they are written in an argumentative, condescending, provoking tone. They are also labeled strong if they contain profanity, verbally attack the opposed opinion or other authors who express it, or use generalized statements about the group of people that have the same or opposite opinion.

\paragraph{Weak.} Statements are annotated with a weak intensity when the author seems to be open or willing to engage in a discussion, sounds less dogmatic, is receptive to another person's opinion or is interested in another’s perspective while expressing their own stance.

\subsection{Additional Labels}\label{sec:additional_labels}

\paragraph{Contextual information requirement.} In most cases the annotated content was part of a conversational thread, therefore we provided the annotators with two types of contextual information. For every annotation, the annotators could optionally view the top-level post in the thread and/or the parent posts that are linked to the target in the hierarchy of the thread, thus every annotation is followed by a label of whether such information was needed. In the guidelines we specifically instructed the annotators to view the context only when they couldn't infer the stance from the standalone text. We release the IDs of the posts that were considered as contextual information regardless of whether they were used or not.

\paragraph{Stance explicitness.} For every annotation, we provide a label that indicates whether the stance of the author was explicitly stated or implied within the text. 

\paragraph{Expression of sarcasm.} Following the work of \newcite{DBLP:conf/evalita/CignarellaFBBPR18} who observed a relationship between texts containing hate speech and irony, we also provide a label for sarcastic content for the posts that contain a stance in order to investigate a possible relationship between stance intensity and sarcasm. By ``sarcastic'' we define any statement that contains sarcasm, irony, humor, wordplay, puns, or any similar figurative language.

\paragraph{Uncertainty of assigned label.} Several sociopolitical issues demand a broader spectrum of knowledge regarding the rules and regulations that are implemented in specific states or countries. Taking into account the annotators' possible unfamiliarity concerning such issues, we introduced a checkbox where the annotators could indicate they are uncertain of their annotation (only in cases where they assigned a stance). However this option  was selected sparingly, i.e. less than 3\% of all annotations.
 
\subsection{Annotation Phases} \label{sec:Annotation_phases}

The dataset was manually annotated by expert annotators and non-experts (NE) with Mechanical Turk. In the first phase, three expert annotators annotated a small subset of posts. The expert annotators were trained on a small subset of posts, where they were able to ask questions and discuss opinions in case of disagreement. In order to measure the performance of the non-expert annotators (phase 1 NE), the posts with at least two out of three expert agreements were used as a control subset. In addition to only accepting the work of NE annotators with high performance in the control subset, we introduced a relatively demanding qualification test to ensure that the NE annotators have a comprehensive understanding of the guidelines. Furthermore, the NE annotators should fill several additional criteria\footnote{Specifically, the NE annotators should have an acceptance rate of 95\%, 1000 approved hits and they should be located either in the US or in the UK in order to ensure that their knowledge of English is sufficient to understand the statements}. Despite these requirements, we observed relatively low values of inter-annotator agreement between the non-experts. In the second phase, around 650 posts were annotated from at least one expert and after computing the pairwise agreements between each NE and the experts, we proceeded to select the top performing annotators to annotate the rest of the dataset (phase 2 NE). 

\section{The Dataset}

\subsection{Data Collection and Filtering}

We built the dataset such that it contains multiple posts by an individual author expressing stances on a specific topic. For each user, we collected statements that were posted on average over the course of 6 months and at most within 15 months. The posts were collected from April 2019 to July 2020. We chose Reddit as our source of data since it provides: (a) rich content, due to the fact that there is no word limit, (b) a clear relationship between the text and the target topic, since users post within a subreddit, and (c) anonymity to some degree. Considering that we wish to observe possible opinion fluctuations, anonymity is extremely important since it mitigates the problem of the spiral of silence effect \cite{https://doi.org/10.1111/j.1460-2466.1974.tb00367.x} and therefore users can be truthful about their beliefs \cite{Meyer16}. Previous work \cite{doi:10.1177/0093650217745429} showed that the controversiality of the topic is one of the main drivers of opinion formation. Since we aimed to observe possible fluctuations in opinion we carefully selected a range of topics that are controversial enough to raise a debate and were related mostly to the users' political identity where we could possibly observe a change in stance \cite{Drummond9587}. Our dataset includes the following topics: (1) abortion, (2) feminism, (3) brexit, (4) veganism/animal rights, (5) guns, (6) nuclear energy, (7) capitalism, and (8) climate change. For each topic, we collected a set of subreddits that encourage debate. To minimize the annotation of irrelevant posts, we applied a binary Linear SVC to provide us a likelihood of a stance being expressed. The classifier was trained on the expert annotations obtained in the first phase of the study. For each user, we selected a set of posts that most likely contained a stance in order to investigate possible opinion fluctuations through time, and annotated all the posts within a set of threads which contained a sufficient amount of posts per user.

\begin{table}[t]
    \centering
    \small
    \begin{tabular}{p{0.75\linewidth}p{0.15\linewidth}}
    \toprule
                                 Attribute & Statistic \\
    \midrule
Total number of annotated posts &      3526 \\
Total number of annotated posts by experts &       673 \\
Total number of annotated authors &       638 \\
Average number of posts per author &      5.53 \\ 
Number of annotated threads trees &       113 \\
Number of annotated thread branches &       547 \\
Number of posts in annotated thread trees &      1752 \\
    \bottomrule
    \end{tabular}
    \caption{General statistics about the annotated dataset.}
    \label{tab:dataset_statistics}
\end{table}

\begin{figure}[h]
\begin{center}
\includegraphics[width=\linewidth]{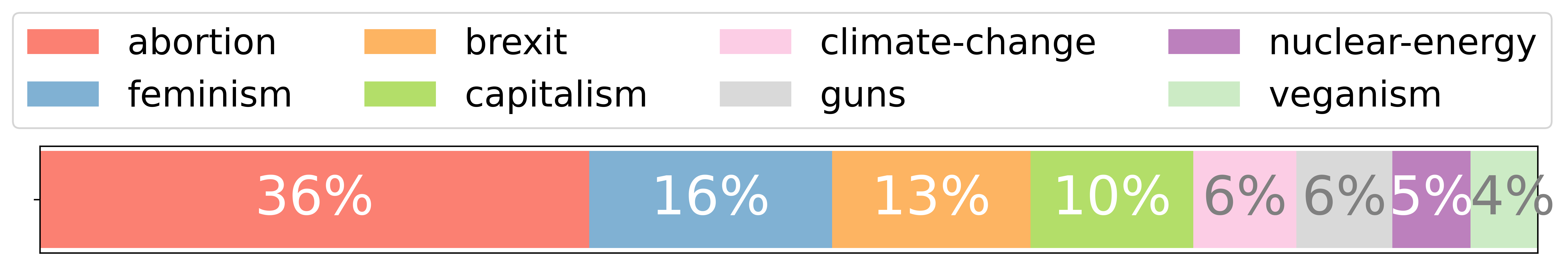} 
\caption{Topic distribution in the dataset}
\label{fig:overall_topic_distribution}
\end{center}
\end{figure}

\subsection{Dataset Statistics}

\paragraph{General statistics.}
We present general statistics about the resulting annotated dataset in Table~\ref{tab:dataset_statistics}. The threads in Reddit have a tree-like structure. When a user replies to another, the different threads of comments become branches of the tree. We show the overall topic distribution in the dataset in Figure~\ref{fig:overall_topic_distribution}. As can be seen, the abortion discussion is the most prominent, making up one third of the dataset, which is due to the high activity in a small number of subreddits.

\begin{figure}[h]
\begin{center}
\includegraphics[width=\linewidth]{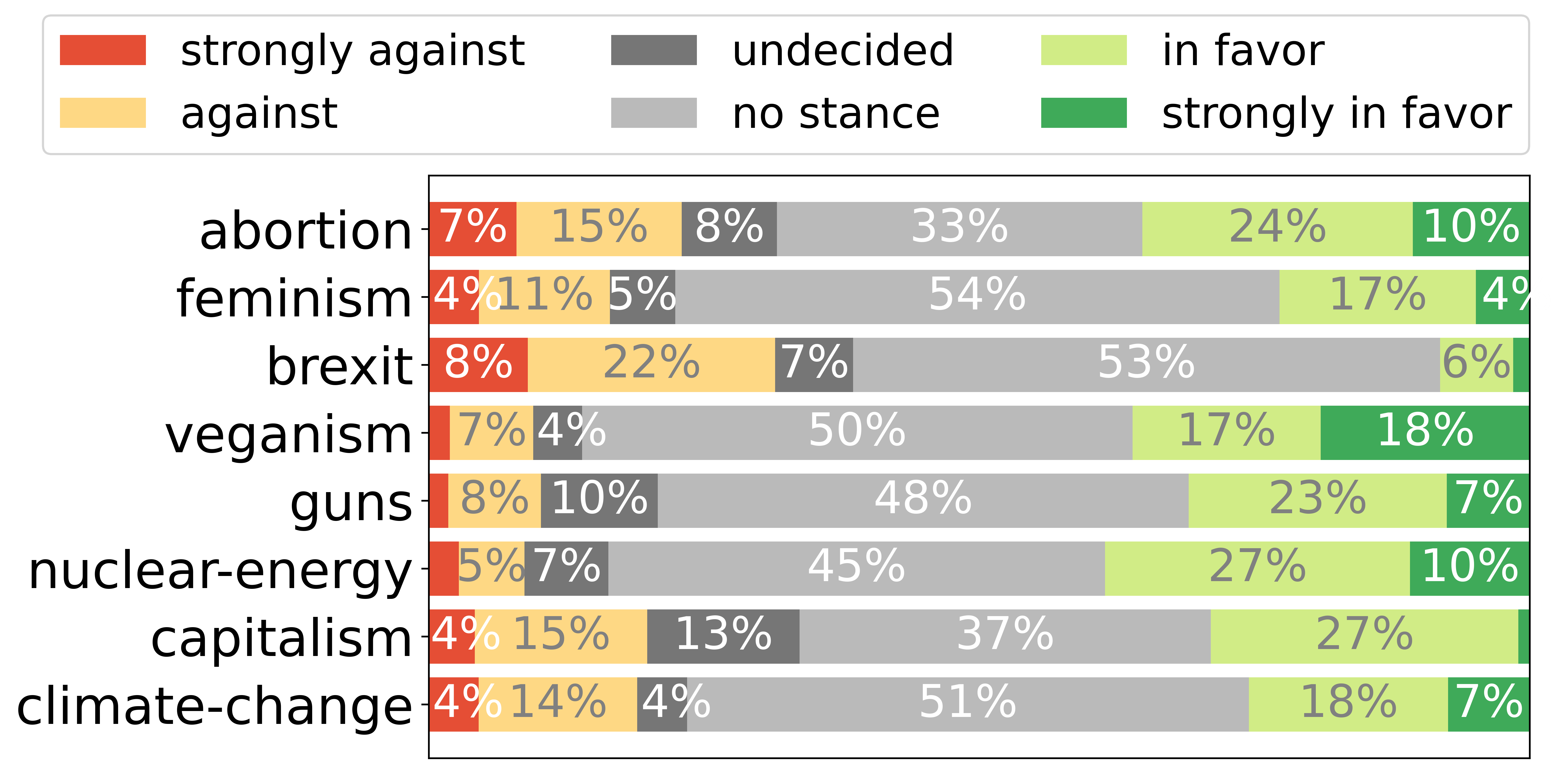} 
\caption{Stance distribution per topic}
\label{fig:stance_distr_per_topic}
\end{center}
\end{figure}

\paragraph{Stance polarity and intensity.} 
The final annotation labels are decided by the majority vote between at least three annotators. We introduce the \textit{undecided} label for the cases where the votes were equally distributed to each polarity and the non inferrable category.
The distributions of stances over the annotated posts of each topic are shown in Figure~\ref{fig:stance_distr_per_topic}. We observe that for all topics most posts do not contain a stance. With the exception of Brexit, the polarities in the rest of the topics tend to lean toward \textit{favor} rather than \textit{against}, which could be biased because of the identified activity on the topics probably relates to an affinity toward the issue. Interestingly, with the exception of veganism, when a stance is expressed, no matter the polarity, the intensity  is mostly weak. 

\paragraph{Contextual information requirement.} We asked the annotators whether the contextual posts were needed for the annotation. In ~42\% of the cases, the annotation was made solely based on the target post's content. For the rest ~58\% of the cases, the annotators selected to view the top-level post ~90\% of the times, the parent posts ~62\% and both top-level and parent posts ~52.54\%. 
Reading the context posts was not rewarded with an additional financial incentive.

\paragraph{Stance explicitness.} Of the annotations that contained stances, only 9.48\% of the stances were annotated as \textit{explicitly expressed} and the remaining 90.52\% were annotated as \textit{implied within the text}. The skew toward implicit expression is reasonable given the loose conversational structure of the discussions. 

\begin{figure}[h]
\begin{center}
\includegraphics[width=\linewidth]{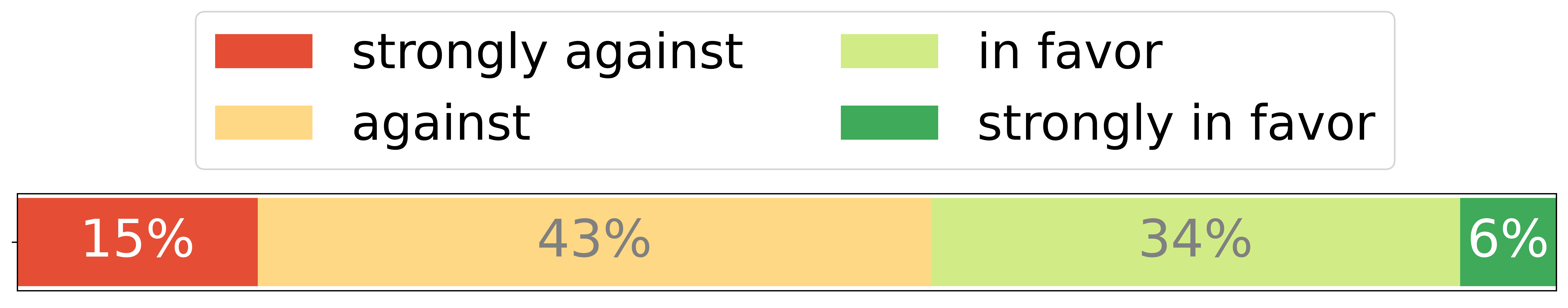} 

\caption{Stance distribution in sarcastic comments}
\label{fig:sarcastic}
\end{center}
\end{figure}

\paragraph{Expression of sarcasm.} Only 3.2\% of the total number of posts contained sarcastic remarks, which is reasonable given that we were not aiming for sarcastic content. Despite observing sarcasm in all topics, the topics with the most sarcastic comments were \textit{abortion} and \textit{Brexit}. Interestingly, the rate of sarcastic posts in Brexit is about double the rate for abortion.
This result follows previous work \cite{10.1525/si.1994.17.1.51,doi:10.1080/14735784.2018.1496843} on the use of sarcasm in politics, which showed that sarcasm is used extensively in political discussions, where users express their frustration and anger through political satire.
Figure~\ref{fig:sarcastic} depicts the stance distribution in sarcastic comments, showing a skew towards the against polarity. We would expect a prevalence of the stronger intensity in sarcastic comments, however the distribution of stance intensity in the sarcastic comments follows that of the whole dataset.

\subsection{Annotator's Statistics}

\begin{figure}[h]%
    \centering
    \subfloat[Annotators' age groups]{{\includegraphics[width=\linewidth]{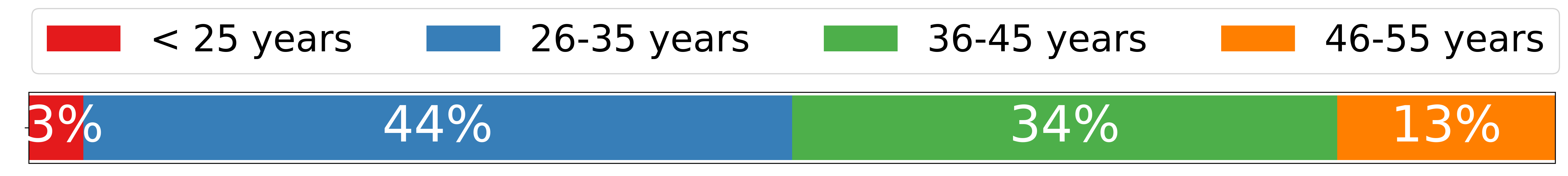} }}%
    
    \subfloat[Annotators' own stances towards each topic]{{\includegraphics[width=\linewidth]{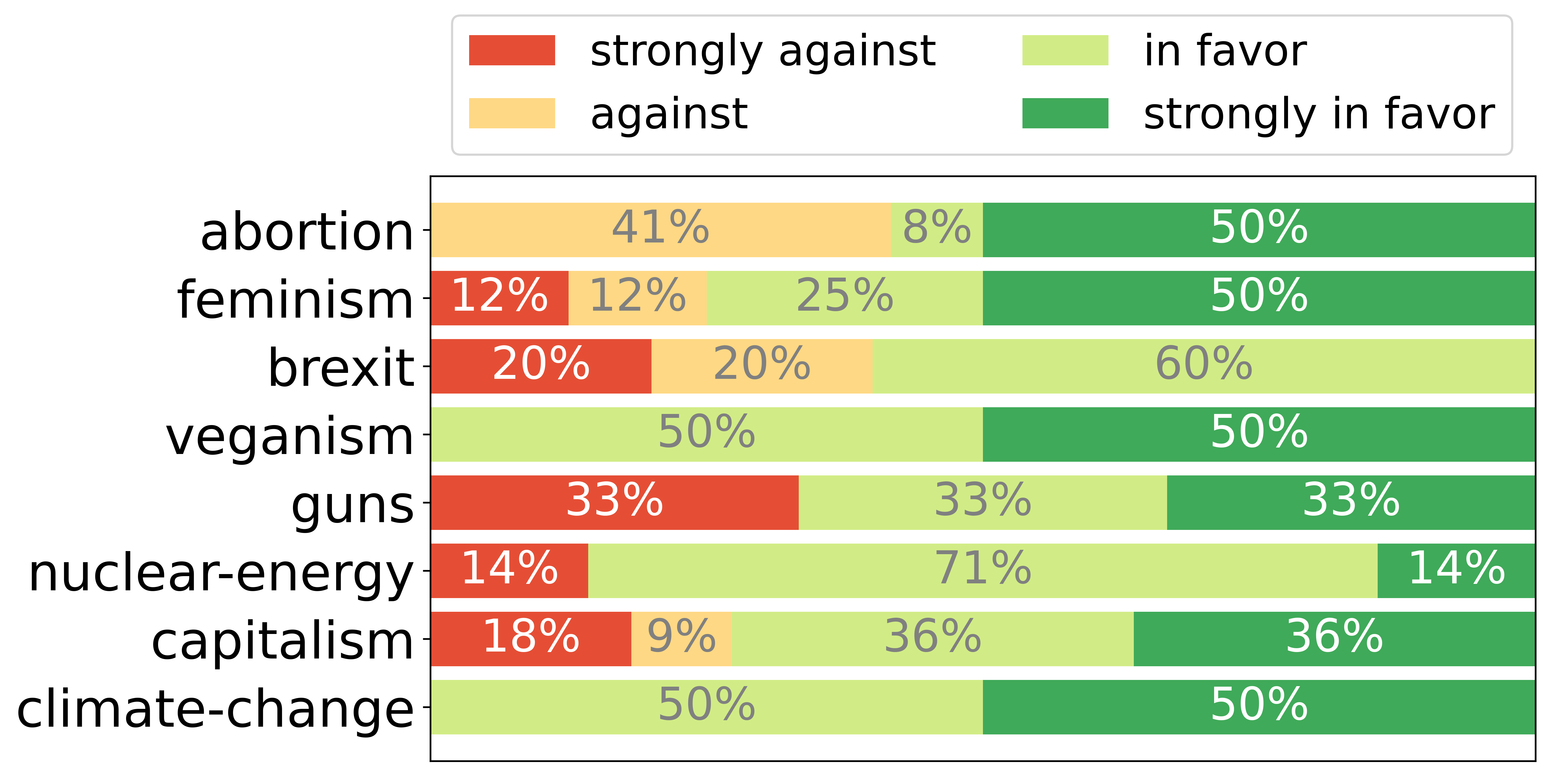} }}%
    \caption{}%
    \label{fig:annotator_stance_and_demographics}%
\end{figure}

\paragraph{Annotator demographics.} We surveyed the annotators about their age and gender demographics and their own stances on each topic. 49\% of the annotators were men, 49\% were women and 2\% were non-binary. We show the rest of the survey results in Figure~\ref{fig:annotator_stance_and_demographics}. There is an obvious prevalence of the favor polarity rather than against. Also, there is a more obvious presence of the strong intensity in the annotator's demographics compared with the resulting dataset; the difference seems reasonable since the annotators' stances were obtained entirely differently, by explicitly asking them, rather than inferring them from text. We computed the cosine similarity between the annotators' personal stances and the average assigned stances and we concluded that the annotators' own biases are not consistently reflected in their annotations.

\begin{figure}[h]
\begin{center}
\includegraphics[width=\linewidth]{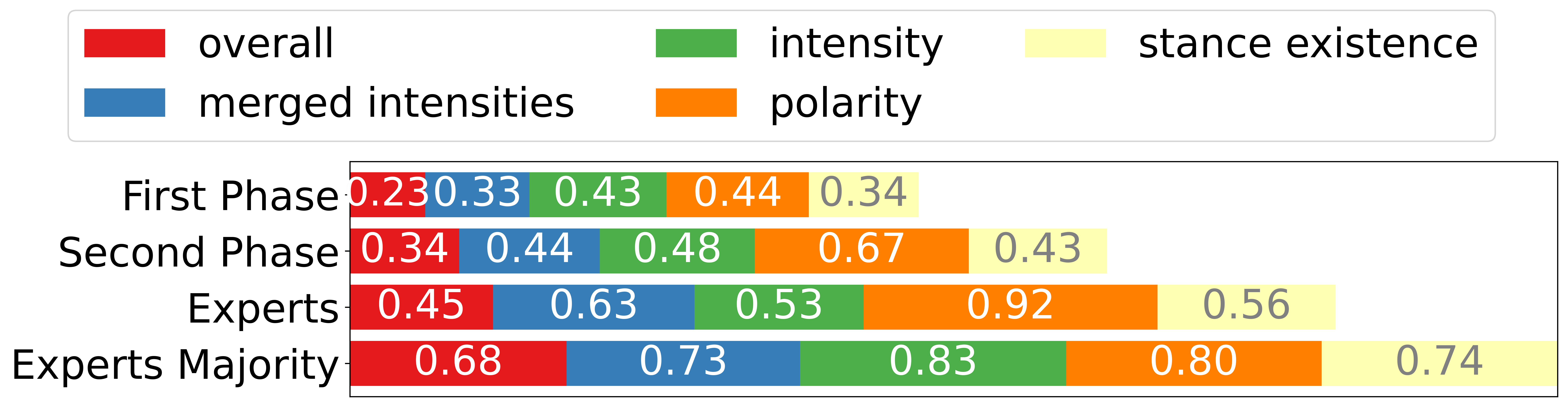} 
\caption{Comparison of inter-annotator agreements between groups of annotators.}
\caption*{This figure shows the inter-annotator agreement between non-experts for the first and second phases, the inter-annotator agreement between the experts and the inter-annotator agreement between the experts and the majority vote of the non-experts}
\label{fig:agreement_by_annotator_group}
\end{center}
\end{figure}

\begin{figure}[h]
\begin{center}
\includegraphics[width=\linewidth]{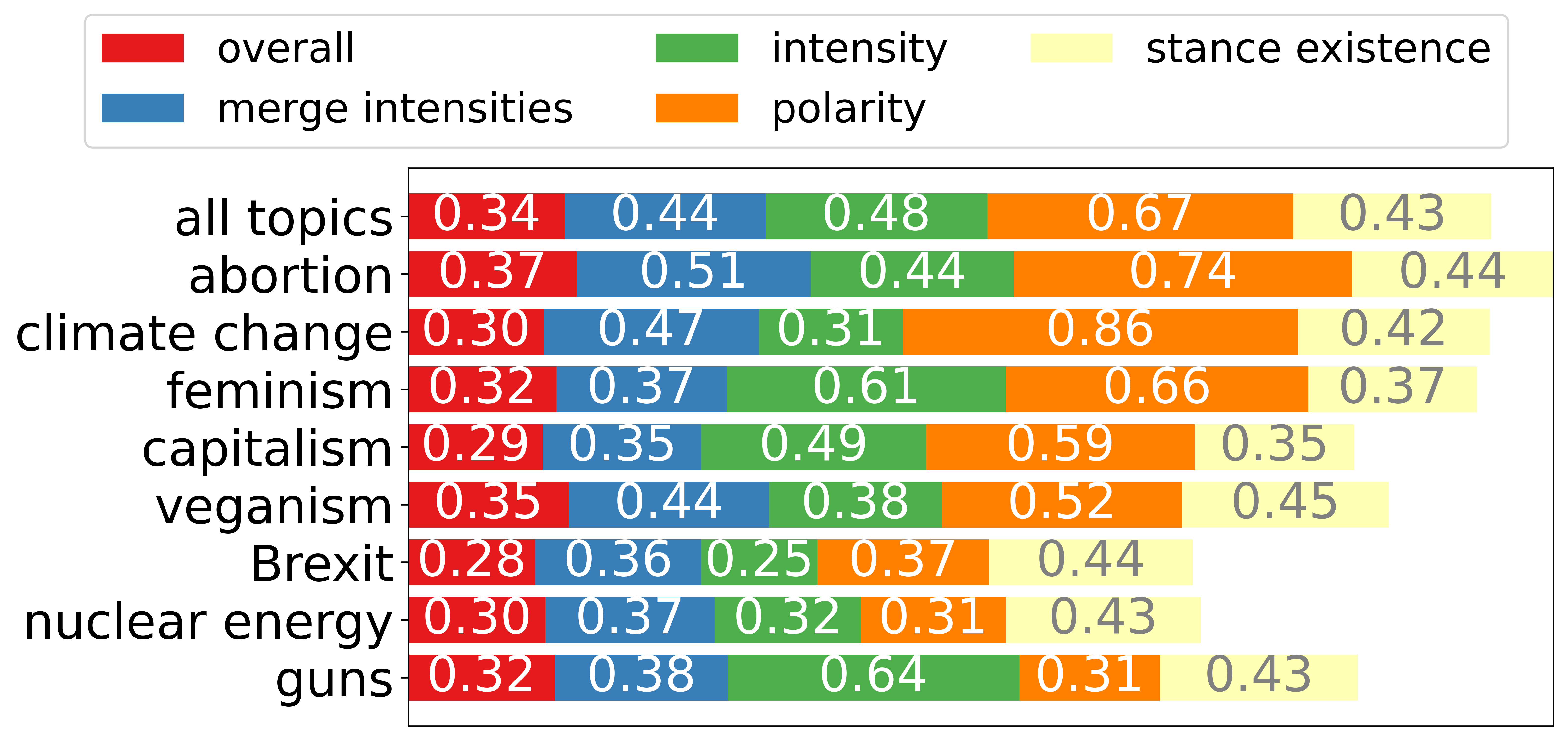} 
\caption{Inter-annotator agreements per topic.}
\label{fig:all_agreements_per_topic}
\end{center}
\end{figure}

\paragraph{Inter-annotator agreements.} 
We compute inter-annotator agreement with Fleiss' Kappa on different aspects of the annotations, displayed in Figures~\ref{fig:agreement_by_annotator_group} and \ref{fig:all_agreements_per_topic}, specifically on the overall five categories (\textit{overall}), the ability to identify the stance without considering the intensity by merging the weak and strong intensities (\textit{merged intensities}) thus resulting in three categories, and the agreement on whether there is an inferrable stance or not (\textit{stance existence}). For the cases where the stance was inferrable, we calculated the agreement on the stance intensity (\textit{intensity}) and the stance polarity (\textit{polarity}). 

Figure~\ref{fig:agreement_by_annotator_group} shows a comparison of inter-annotator agreements between groups the NE annotators from the first phase of the project (phase 1 NE), the selected NE annotators from the second phase (phase 2 NE) and the  experts. Following the work of \newcite{10.1145/1743384.1743478}, we computed the pairwise agreements between the majority vote of the phase 2 NE and the expert annotators in order to demonstrate the quality of the resulting annotations.

When comparing the agreements among the three different annotator groups, we can see that the agreements were best among the expert group, followed by the phase 2 NE, and lastly among the phase 1 NE. The stark differences highlight the difficulty of the task, that such a task is best performed by a group of experts. The expert group attained a moderate overall agreement and a substantial agreement when merging the intensities. However, we only have expert annotations for a small portion of the data, given the difficulty of scaling expert annotations. As using experts is the most costly approach, the recruiting method we utilized as described in Section~\ref{sec:Annotation_phases}, offered significantly better agreements balancing costs and performance. In addition, we show that the resulting annotations obtained from the majority vote are of comparable quality to the annotations of the experts. More specifically, we obtain a substantial agreement on overall categories and when the intensities are merged, an almost perfect agreement on both the extremity and the polarity. 

Since the final dataset contains the phase 2 NE annotations, we  break this analysis down further in Figure~\ref{fig:all_agreements_per_topic} by analyzing the phase 2 NE agreements per topic.

We observe low overall agreement in specific topics, namely Brexit, nuclear energy and capitalism. This could be explained by the fact that reading statements about Brexit requires broad knowledge about the political landscape of the UK as well as some knowledge of economic theory. In the case of capitalism, we observed that there is a heavy use of terminology regarding various economic and political systems, which might be unfamiliar to the non-expert annotators. The same issue appears in the nuclear energy topic, in which several posts contain highly technical and scientific terminology.  On the other hand, topics such as abortion and veganism show higher overall agreement since the posts are about everyday issues. 

\section{Case Studies}

This section presents two case studies on the dataset. The first analyzes users based on the stability of their opinions. Second, we tested simple baselines to establish benchmarks on several different types of stance detection tasks.

\subsection{User Stance Entropy}

We computed the user's entropy as an indicator of opinion change, more specifically, high entropy users tend to oscillate between different stances, while users with low entropy have a stable opinion polarity and might only change the intensity of their stance. To create each group, we compute the quartiles of the entropies, assigning users in the bottom quartile to \textit{low} and the top quartile to \textit{high}. 

\paragraph{Opinion fluctuation.}
We explore the users' opinion fluctuations through their posting history, as well as within a discussion thread. We wish to show whether the opinion change happens over time or instantly after a debate with users of the opposite opinion.

\begin{figure}[h]%
    \centering
    \subfloat{{\includegraphics[width=\linewidth]{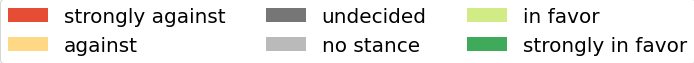}  }}%
    \vspace{-1em}
    \subfloat{{\includegraphics[width=\linewidth]{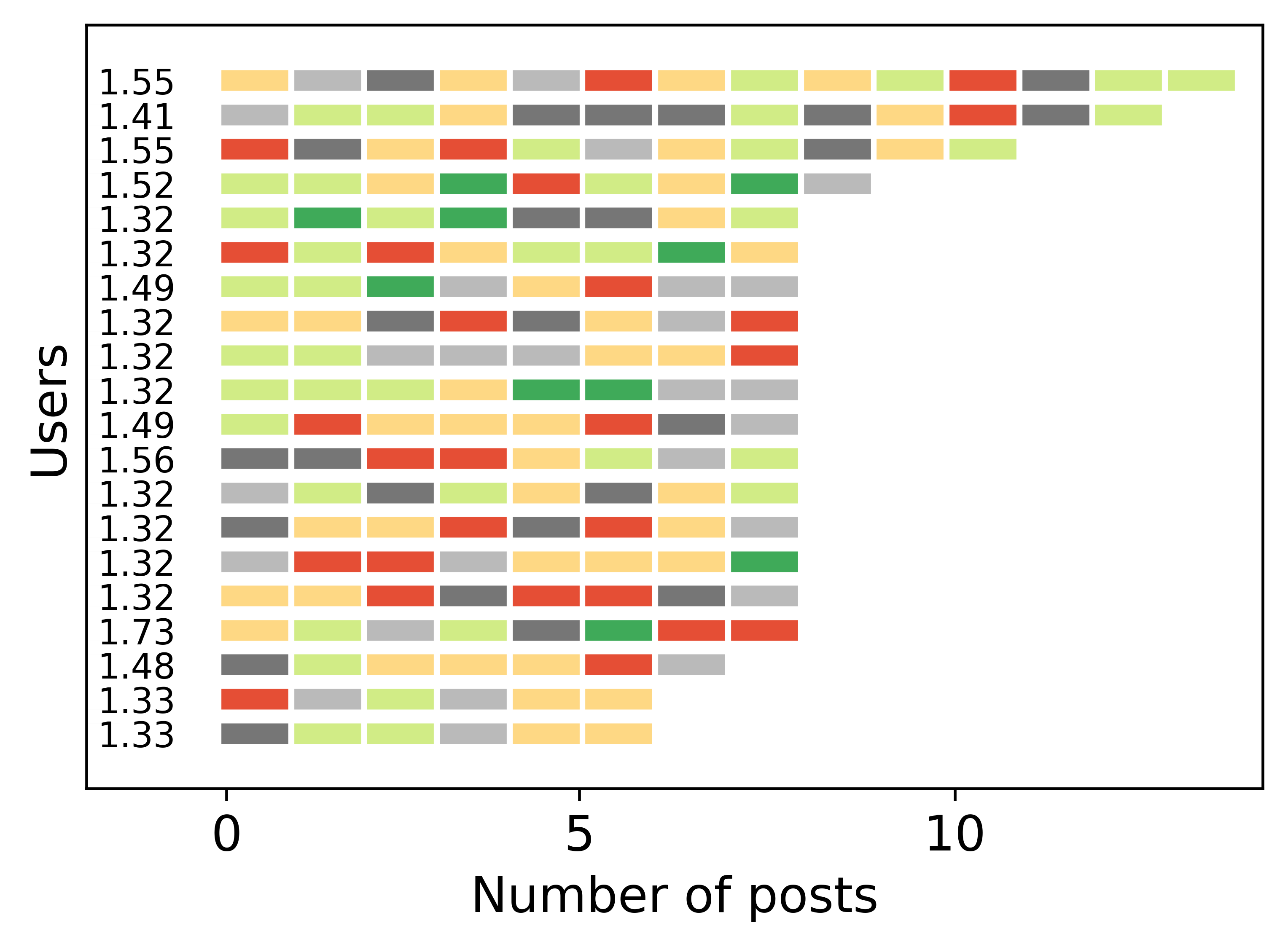} }}%
    \qquad
    \caption{Evolution of the users' stance through time. Each row represents a distinct user and, each block represents a post and that user's stance towards abortion. The y axis shows each user's entropy. }%
    \label{fig:user_stance_evo_over_time}%
\end{figure}

Figure~\ref{fig:user_stance_evo_over_time} shows the evolution of the opinionated users' stances through time for the topic abortion. We can see from the Figure that in most cases, users change the polarity of their stance, however they rarely show strong intensities for both polarities. After investigating this behavior, we observe vacillating attitudes in users with posting periods of over one month. Similar patterns in opinion change are present for the rest of the topics. That means that in most cases the users' opinion oscillates between polarities and there is not only one specific point in time where there is a switch in stance from one pole to another. Table~\ref{tab:opinion_change} shows a selected sample of annotated posts from one specific user and how this user's opinion changes over time regarding abortion. Note that the text has been slightly modified to preserve anonymity.

\begin{table}[t]
    \centering
    \small
    \begin{tabular}{p{\linewidth}}
         \textit{``(...) The ONLY thing we (pro-lifers) are concerned with is ending the killing of unborn babies.''} \\
         \textbf{Stance:  {\color{Red}Strongly against} abortion}\\        
         \textit{``I could consider an abortion to save a woman's life. (...)'' }\\
         \textbf{Stance: {\color{ForestGreen}Weakly favors} abortion} \\
        \midrule
         \textit{``After looking into this further I found out that (...) in some cases an abortion is actually more threatening to the woman than delivering the baby.''} \\
         \textbf{Stance: {\color{YellowOrange}Weakly against} abortion}\\
        \midrule
         \textit{``Ectopic pregnancy is a whole different world and I'm sure abortion usually is necessary in those cases. (...) Therefore, abortion is not life-threatening to women.''} \\
         \textbf{Stance:  {\color{ForestGreen}Weakly favors} abortion}\\
         \textit{``A human fetus is as human as a toddler. It has a right to life.''} \\
         \textbf{Stance:  {\color{YellowOrange}Weakly against} abortion}\\
        \cdashline{1-1}
    \end{tabular}
    \caption{Example of an individual's opinion change towards abortion.}
    \label{tab:opinion_change}
\end{table}

\begin{figure}[h]
\begin{center}
\includegraphics[width=\linewidth]{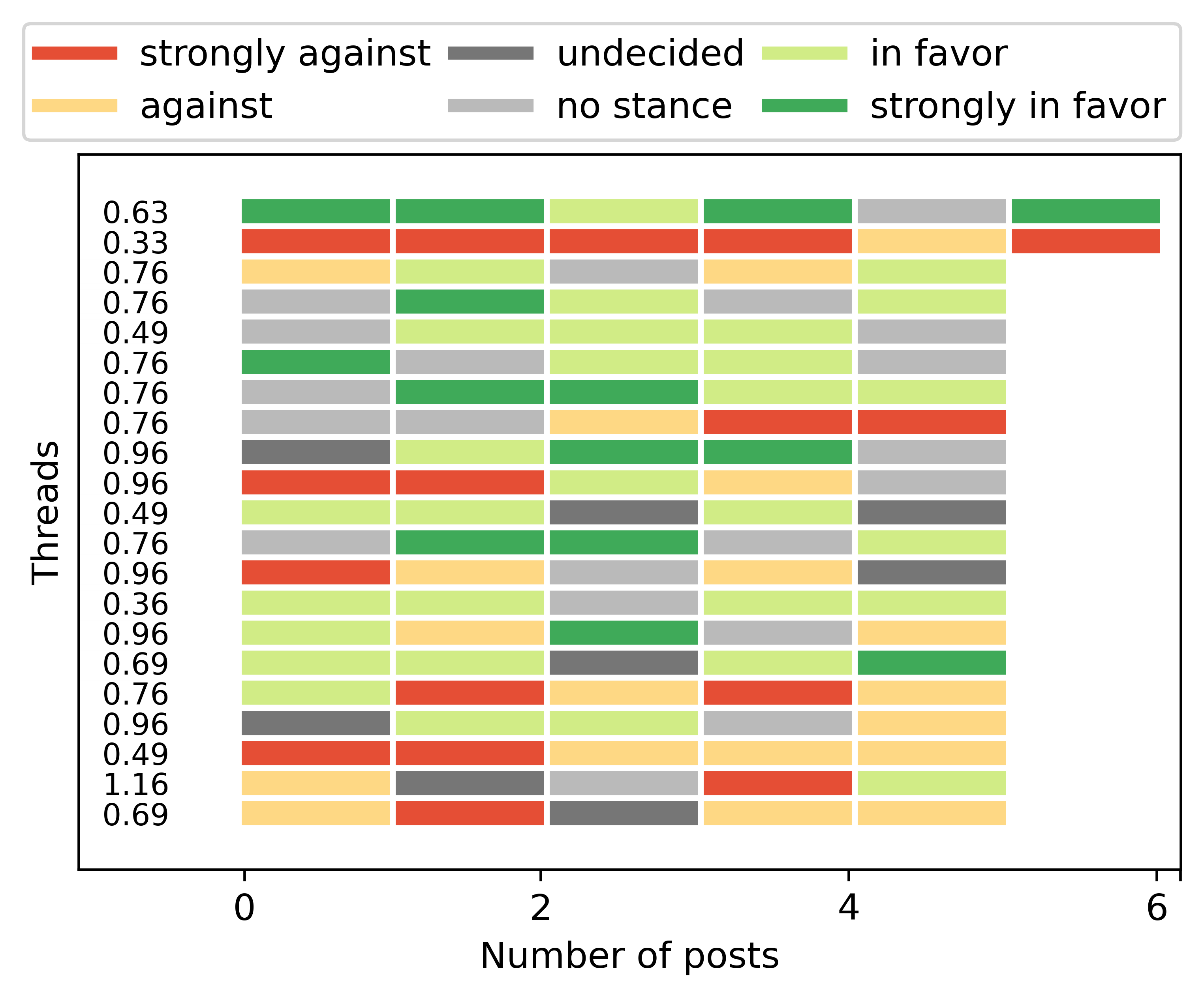} 
\caption{Evolution of the users' stance within a thread. Each row represents a distinct user and, each block represents a post and that user's stance towards a topic. On the y axis we show each user's entropy.}
\label{fig:user_stance_evolution_thread}
\end{center}
\end{figure}

To observe the evolution of the users' opinion within a conversation, we selected the top 20 users with the highest entropy. Figure~\ref{fig:user_stance_evolution_thread} shows the evolution of these users' stances within a thread against various topics. We can see that in most cases the users oscillate their stance intensity and rarely their polarity, leading us to the conclusion that opinion change happens gradually over time or at least it is not immediately expressed within the conversation. 

Since there is no obvious pattern in opinion change, we need to determine the real reason behind these opinion fluctuations. Is it true that the user can understand the opposite side in selected cases (such as the example in Table~\ref{tab:opinion_change}), is it random, or maybe a result of noisy annotations? The consistency of these changes is still a subject to investigation, as well as the conditions under which a user changes their mind. All in all, this dataset enables the investigation of such questions.

\paragraph{Language usage.}
We compare the language usage of high-entropy users to low-entropy and analyze their language usage using the LIWC (Linguistic Inquiry and Word Count) lexicon~\cite{pennebakerlinguistic}. This methodology quantifies different categories of words in the text such as the frequency of causal, emotional, and cognitive words. Table~\ref{tab:liwc_overall} shows the LIWC categories that are most highly related to each user group by positive pointwise  mutual  information (ppmi). We focus our discussion on the \textit{health} category, which is particularly interesting in how it relates to stance entropy and topics.

\begin{table}[t]
    \centering
    \small
    \begin{tabular}{cc}
      \begin{tabular}{lr}
    \toprule
      category & ppmi (high) \\
    \midrule
    family &         .456 \\
     death &         .426 \\
    health &          .390 \\
    sexual &         .388 \\
    female &         .255 \\
    \bottomrule
    \end{tabular} & 
    
    \begin{tabular}{lr}
    \toprule
     category & ppmi (low) \\
    \midrule
      filler &        .416 \\
      ingest &        .271 \\
         home &        .262 \\
      leisure &        .259 \\
         feel &        .138 \\
    \bottomrule
    \end{tabular}
    \\
    \end{tabular}
    \caption{Top LIWC five categories based on positive pointwise mutual information for high-entropy users (left) and low-entropy users (right).}
    \label{tab:liwc_overall}
\end{table}

The health category includes terms such as \textit{abortion, life, pregnancy, clinic, medical,} and \textit{alive}. Considering all text by each group of users, there is higher ppmi with high entropy users. However, this is a particularly interesting feature because it depends on the topic, with \textit{climate change}, \textit{guns}, and \textit{veganism} showing higher ppmi with low entropy users. To understand these differences, we further analyze posts with health features, considering the stances and stance entropy of their authors by topic and observe qualitative examples.

For \textit{climate change} and \textit{guns}, the low entropy stances of the authors lean toward a particular pole, respectively \textit{against} the notion that climate change is caused by humans, and \textit{in favor} of guns. In contrast, the low entropy stances are evenly split across the poles for \textit{veganism}, suggesting users who express their stance on veganism in relation to health are firm on their stances no matter the polarity.

Next, we qualitatively observe these samples. We observed that for \textit{climate change} and \textit{guns} the feature is overly driven by the word ``exercise" which is used in the context of \textit{exercising one's rights.} The veganism topic offers a more interesting case study, not only because the health features are accurately driven by health contexts, but also, and mainly, because of the even split across poles. Particularly, we observed that the health contexts differ across the opposing polarities.  Among \textit{favor} posts on veganism, health discussions are primarily about diet and environmental issues. For example, one user states, ``health can quite easily be achieved sans animal products." Among \textit{against} posts on veganism, health contexts are primarily about animal testing research. For example, one user says, ``If I'm taking a drug for the heart, I wanna know if there are any side effects on the other organs. Animal models, especially mammals (rats, mice) are the next best thing we have to testing humans."

\subsection{Stance Detection}

We experiment with simple baselines on various stance detection tasks that can be performed on our dataset, and offer a brief discussion of the results. We performed the following classification tasks:

\begin{enumerate}
    \item Has stance: binary classification of whether the post has a stance or the stance is not inferrable
    \item Intensity: binary classification of whether a post's stance is weak or strong regardless of its polarity
    \item Polarity: three-class classification of the polarity (against/favor) or stance is not inferrable
    \item Fine-grained: four-class classification of the inferrable stances with intensities (i.e., strongly against, against, favor, strongly favor)
    \item All: five-class classification of the fine-grained stances, including stance not inferrable and excluding the posts from the ``undecided'' class.
    \item All-maj: Same as ``All'' but taking into account only the posts with majority agreement (at least two of three votes)
\end{enumerate}

We evaluated several different n-gram models (up to trigrams) with Naive Bayes, Linear SVC, and Logistic Regression classifiers and a bert-based approach in a 10-fold setup. As the main focus of this work was the development of the dataset, these models are meant to be simple and serve as benchmarks; we did not perform extensive experiments with varied parameters or feature engineering, but encouraged further experimentation for future work.

Table~\ref{tab:results_all_tasks} shows the weighted F1 scores of the best performing models for each classification task, alongside the weighted F1's of random prediction baseline based on the label distribution of the training set, and of a majority class assignment baseline. All best-performing models were either Logistic Regression or Naive Bayes on unigram features, which all outperformed the random baselines significantly.~\footnote{$p < .0001$ by a bootstrap resampling significance test}

Though the models are capable of outperforming the majority baselines, their performances are low within reason, as they reflect the difficulty of the task experienced by the human annotators. If we consider, for instance, how much the model for the \textit{polarity} three-class task improves over the majority class baseline broken down by topic, the topics with highest agreement also have the models with the most significant improvement. That is, the \textit{abortion} and \textit{climate change} topics have both the highest agreement and most significant improvement over the majority class, while \textit{capitalism} and \textit{nuclear energy} have much lower agreement and the model shows no significant improvement.

\begin{table}[t]
    \centering
    \footnotesize
\begin{tabular}{lp{1cm}p{1.2cm}p{1.2cm}}
\toprule
            Task (\# classes) &  Top Model &  Random Baseline &  Majority Baseline \\
\midrule
            Has stance (2) &        0.660 &         0.510 &                  0.365 \\
            Intensity (2) &    0.652 &          0.594 &                  0.609 \\
            Polarity (3)  &        0.495 &           0.344 &                  0.198 \\
            Fine-grained (4) &        0.373 &                 0.307 &                  0.258 \\
            All (5) &        0.341 &                 0.230 &                  0.124 \\
            All-maj (5) &        0.340 &           0.211 &                  0.124 \\
\bottomrule
\end{tabular}
    \caption{Top performing models by weighted F1 score for each classification task.}
    \label{tab:results_all_tasks}
\end{table}

\section{Conclusions and Future Work}

We introduce an innovative dataset for detecting subtle opinion changes and fine-grained stances. The dataset contains a sufficient amount of stance polarities and intensities per user on various sociopolitical topics over time but also within entire conversational threads. We recruited reliable non-expert annotators based on their agreement with the experts and show that, through the majority vote the resulting set of annotations is of comparable quality to that of the experts. We observe low inter-annotator agreement in topics with highly technical and scientific terminology, indicating that broader knowledge about a specific topic is necessary to obtain high quality annotations.

We analyze the opinion fluctuations of users with vacillating attitudes. When analyzing these fluctuations over time, we observe that users change the polarity of their stance rather frequently, however most users express strong intensities for only one of the two poles. When analyzing these fluctuations within conversational threads, we observe that there is usually only a change in intensity and rarely in the polarity, leading us to the conclusion that opinion change is expressed over long periods of time.  Through a case study analysis we observe that specific word categories are more prevalent in the vocabularies of vacillating users. Finally, we provide baseline results for different setups, showing that this dataset offers a range of challenging tasks, while leading to new, interesting research paths.

Our dataset addresses the need for temporal data to improve stance detection models that utilize the user history \cite{Monti_2020,10.5555/3157096.3157141,wrap116715}. In addition, the users’ historical posts can be used to model their temporal behavior in order to predict subtle opinion shifts. Furthermore, future work could further investigate relationships between intensity persistence and temporal aspects of user behaviors. Such work could consider relations to real-world events, forum engagement activity, and linguistic changes. There are other relevant linguistic aspects to explore, related to the field of argumentation, such as whether receptive language \cite{Sobkowicz2012DiscreteMO} relates to opinion or intensity changes, or linguistic style matching in interactions between users with shared or opposite opinions. In addition, one could also investigate the potential relationship between the expression of various emotions and stance intensity. In conclusion, this dataset enables the investigation of various unexplored research directions by raising interesting, new questions about the user's behavior and intersecting the field of sociopolitical sciences with natural language processing.

\section*{Appendix}
\subsection*{Ethical Considerations}

The ability to automatically predict the personal behavior of online users in order to improve natural language classification algorithms can raise particular, sometimes non-obvious ethical concerns. Use of any user data for personalization shall be anonymous and non-identifiable, and limited to the given purpose, no individual posts shall be republished \cite{hewson2013ethics}. For this reason, only the raw content of the posts is publicly released in our dataset without any connection to the post-ids or the user handles connected to the posts. All user data is kept separately on protected servers, linked to the raw text and network data only through anonymous IDs. In addition, while deploying the studies in Amazon Mechanical Turk we made sure that possible user handles that were mentioned in the text were not disclosed to the annotators. Researchers are advised to take account of users’ expectations \cite{williams2017towards,shilton2016we,townsend2016social} when collecting public data such as Reddit. 

In addition, the dynamic analysis and forecasting of opinion can be used to identify easily swayed individuals which can be used for harmful purposes. Furthermore, it is important we are mindful of the fact that certain groups with no access or difficulty navigating the Internet are underrepresented in this data and the stance towards specific sociopolitical issues is not the general consensus of the public but the voice of a mere percentage of a localized group of people.

\subsection*{Acknowledgements}
This work has been supported by the German Federal Ministry of Education and Research (BMBF) as a part of the Junior AI Scientists program under the reference 01-S20060.

\section{Bibliographical References}\label{reference}


\begin{thebibliography}{}

    \bibitem[\protect\citename{Acemoglu and
      Ozdaglar}2011]{RePEc:spr:dyngam:v:1:y:2011:i:1:p:3-49}
    Acemoglu, D. and Ozdaglar, A.
    \newblock (2011).
    \newblock {Opinion Dynamics and Learning in Social Networks}.
    \newblock {\em Dynamic Games and Applications}, 1(1):3--49, March.
    
    \bibitem[\protect\citename{Brock}2018]{doi:10.1080/14735784.2018.1496843}
    Brock, M.
    \newblock (2018).
    \newblock Political satire and its disruptive potential: irony and cynicism in
      russia and the us.
    \newblock {\em Culture, Theory and Critique}, 59(3):281--298.
    
    \bibitem[\protect\citename{Cignarella \bgroup et al.\egroup
      }2018]{DBLP:conf/evalita/CignarellaFBBPR18}
    Cignarella, A.~T., Frenda, S., Basile, V., Bosco, C., Patti, V., and Rosso, P.
    \newblock (2018).
    \newblock Overview of the {EVALITA} 2018 task on irony detection in italian
      tweets (ironita).
    \newblock In Tommaso Caselli, et~al., editors, {\em Proceedings of the Sixth
      Evaluation Campaign of Natural Language Processing and Speech Tools for
      Italian. Final Workshop {(EVALITA} 2018) co-located with the Fifth Italian
      Conference on Computational Linguistics (CLiC-it 2018), Turin, Italy,
      December 12-13, 2018}, volume 2263 of {\em {CEUR} Workshop Proceedings}.
      CEUR-WS.org.
    
    \bibitem[\protect\citename{Coates \bgroup et al.\egroup
      }2018]{10.5555/3237383.3237857}
    Coates, A., Han, L., and Kleerekoper, A.
    \newblock (2018).
    \newblock A unified framework for opinion dynamics.
    \newblock In {\em Proceedings of the 17th International Conference on
      Autonomous Agents and MultiAgent Systems}, AAMAS '18, page 1079–1086,
      Richland, SC. International Foundation for Autonomous Agents and Multiagent
      Systems.
    
    \bibitem[\protect\citename{De \bgroup et al.\egroup
      }2016]{10.5555/3157096.3157141}
    De, A., Valera, I., Ganguly, N., Bhattacharya, S., and Gomez-Rodriguez, M.
    \newblock (2016).
    \newblock Learning and forecasting opinion dynamics in social networks.
    \newblock In {\em Proceedings of the 30th International Conference on Neural
      Information Processing Systems}, NIPS'16, page 397–405, Red Hook, NY, USA.
      Curran Associates Inc.
    
    \bibitem[\protect\citename{Dowling \bgroup et al.\egroup
      }2020a]{doi:10.1177/1532673X19832543}
    Dowling, C.~M., Henderson, M., and Miller, M.~G.
    \newblock (2020a).
    \newblock Knowledge persists, opinions drift: Learning and opinion change in a
      three-wave panel experiment.
    \newblock {\em American Politics Research}, 48(2):263--274.
    
    \bibitem[\protect\citename{Dowling \bgroup et al.\egroup
      }2020b]{dowling_henderson_miller_2019}
    Dowling, C.~M., Henderson, M., and Miller, M.~G.
    \newblock (2020b).
    \newblock Knowledge persists, opinions drift: Learning and opinion change in a
      three-wave panel experiment.
    \newblock {\em American Politics Research}, 48(2):263--274.
    
    \bibitem[\protect\citename{Drummond and Fischhoff}2017]{Drummond9587}
    Drummond, C. and Fischhoff, B.
    \newblock (2017).
    \newblock Individuals with greater science literacy and education have more
      polarized beliefs on controversial science topics.
    \newblock {\em Proceedings of the National Academy of Sciences},
      114(36):9587--9592.
    
    \bibitem[\protect\citename{Ducharme}1994]{10.1525/si.1994.17.1.51}
    Ducharme, L.~J.
    \newblock (1994).
    \newblock Sarcasm and interactional politics.
    \newblock {\em Symbolic Interaction}, 17(1):51--62.
    
    \bibitem[\protect\citename{Durmus and
      Cardie}2018]{durmus-cardie-2018-exploring}
    Durmus, E. and Cardie, C.
    \newblock (2018).
    \newblock Exploring the role of prior beliefs for argument persuasion.
    \newblock In {\em Proceedings of the 2018 Conference of the North {A}merican
      Chapter of the Association for Computational Linguistics: Human Language
      Technologies, Volume 1 (Long Papers)}, pages 1035--1045, New Orleans,
      Louisiana, June. Association for Computational Linguistics.
    
    \bibitem[\protect\citename{Edwards}1996]{EDWARDS1996}
    Edwards, K;~Smith, E.~E.
    \newblock (1996).
    \newblock A disconfirmation bias in the evaluation of arguments.
    \newblock {\em Journal of personality and social psychology}.
    
    \bibitem[\protect\citename{Flache}2019]{Flache2019}
    Flache, A., (2019).
    \newblock {\em Social Integration in a Diverse Society: Social Complexity
      Models of the Link Between Segregation and Opinion Polarization}, pages
      213--228.
    \newblock Springer International Publishing, Cham.
    
    \bibitem[\protect\citename{Hewson and Buchanan}2013]{hewson2013ethics}
    Hewson, C. and Buchanan, T.
    \newblock (2013).
    \newblock Ethics guidelines for internet-mediated research.
    \newblock The British Psychological Society.
    
    \bibitem[\protect\citename{Jager}2005]{Jager05uniformitybipolarization}
    Jager, W.
    \newblock (2005).
    \newblock Uniformity, bipolarization and pluriformity captured as generic
      stylized behavior with an agent-based simulation model of attitude change.
    \newblock {\em Computational and Mathematical Organization Theory}, page 303.
    
    \bibitem[\protect\citename{{Levow} \bgroup et al.\egroup }2014]{7078580}
    {Levow}, G., {Freeman}, V., {Hrynkevich}, A., {Ostendorf}, M., {Wright}, R.,
      {Chan}, J., {Luan}, Y., and {Tran}, T.
    \newblock (2014).
    \newblock Recognition of stance strength and polarity in spontaneous speech.
    \newblock In {\em 2014 IEEE Spoken Language Technology Workshop (SLT)}, pages
      236--241.
    
    \bibitem[\protect\citename{Lord \bgroup et al.\egroup }1979]{pub.1052753222}
    Lord, C.~G., Ross, L., and Lepper, M.~R.
    \newblock (1979).
    \newblock Biased assimilation and attitude polarization: The effects of prior
      theories on subsequently considered evidence.
    \newblock {\em Journal of Personality and Social Psychology},
      37(11):2098--2109.
    
    \bibitem[\protect\citename{M\"{a}s and
      Flache}2013]{10.1371/journal.pone.0074516}
    M\"{a}s, M. and Flache, A.
    \newblock (2013).
    \newblock Differentiation without distancing. explaining bi-polarization of
      opinions without negative influence.
    \newblock {\em PLOS ONE}, 8(11):1--17, 11.
    
    \bibitem[\protect\citename{M\"{a}s \bgroup et al.\egroup
      }2013]{10.1287/orsc.1120.0767}
    M\"{a}s, M., Flache, A., Tak\'{a}cs, K., and Jehn, K.~A.
    \newblock (2013).
    \newblock In the short term we divide, in the long term we unite: Demographic
      crisscrossing and the effects of faultlines on subgroup polarization.
    \newblock {\em Organization Science}, 24(3):716–736, May.
    
    \bibitem[\protect\citename{Matthes \bgroup et al.\egroup
      }2018]{doi:10.1177/0093650217745429}
    Matthes, J., Knoll, J., and von Sikorski, C.
    \newblock (2018).
    \newblock The spiral of silence revisited: A meta-analysis on the relationship
      between perceptions of opinion support and political opinion expression.
    \newblock {\em Communication Research}, 45(1):3--33.
    
    \bibitem[\protect\citename{Matthes}2014]{10.1093/ijpor/edu032}
    Matthes, J.
    \newblock (2014).
    \newblock {Observing the “Spiral” in the Spiral of Silence}.
    \newblock {\em International Journal of Public Opinion Research},
      27(2):155--176, 10.
    
    \bibitem[\protect\citename{Meyer and Speakman}2016]{Meyer16}
    Meyer, H. and Speakman, B.
    \newblock (2016).
    \newblock Quieting the commenters: The spiral of silence's persistent effect on
      online news forums quieting the commenters: The spiral of silence's
      persistent effect on online news forums.
    \newblock {\em International Symposium on Online Journalism}, 6, 05.
    
    \bibitem[\protect\citename{Mohammad \bgroup et al.\egroup
      }2016a]{mohammad-etal-2016-dataset}
    Mohammad, S., Kiritchenko, S., Sobhani, P., Zhu, X., and Cherry, C.
    \newblock (2016a).
    \newblock A dataset for detecting stance in tweets.
    \newblock In {\em Proceedings of the Tenth International Conference on Language
      Resources and Evaluation ({LREC} 2016)}, pages 3945--3952, Portoro{\v{z}},
      Slovenia, May. European Language Resources Association (ELRA).
    
    \bibitem[\protect\citename{Mohammad \bgroup et al.\egroup
      }2016b]{mohammad-etal-2016-semeval}
    Mohammad, S., Kiritchenko, S., Sobhani, P., Zhu, X., and Cherry, C.
    \newblock (2016b).
    \newblock {S}em{E}val-2016 task 6: Detecting stance in tweets.
    \newblock In {\em Proceedings of the 10th International Workshop on Semantic
      Evaluation ({S}em{E}val-2016)}, pages 31--41, San Diego, California, June.
      Association for Computational Linguistics.
    
    \bibitem[\protect\citename{Monti \bgroup et al.\egroup }2020]{Monti_2020}
    Monti, C., De~Francisci~Morales, G., and Bonchi, F.
    \newblock (2020).
    \newblock Learning opinion dynamics from social traces.
    \newblock {\em Proceedings of the 26th ACM SIGKDD International Conference on
      Knowledge Discovery and Data Mining}, Jul.
    
    \bibitem[\protect\citename{Noelle-Neumann}1974]{https://doi.org/10.1111/j.1460-2466.1974.tb00367.x}
    Noelle-Neumann, E.
    \newblock (1974).
    \newblock The spiral of silence a theory of public opinion.
    \newblock {\em Journal of Communication}, 24(2):43--51.
    
    \bibitem[\protect\citename{Nowak and R\"{u}ger}2010]{10.1145/1743384.1743478}
    Nowak, S. and R\"{u}ger, S.
    \newblock (2010).
    \newblock How reliable are annotations via crowdsourcing: A study about
      inter-annotator agreement for multi-label image annotation.
    \newblock In {\em Proceedings of the International Conference on Multimedia
      Information Retrieval}, MIR '10, page 557–566, New York, NY, USA.
      Association for Computing Machinery.
    
    \bibitem[\protect\citename{Nyhan and Reifler}2010]{Nyhan2010}
    Nyhan, B. and Reifler, J.
    \newblock (2010).
    \newblock When corrections fail: The persistence of political misperceptions.
    \newblock {\em Political Behavior}, 32(2):303--330, June.
    
    \bibitem[\protect\citename{Pennebaker \bgroup et al.\egroup
      }2015]{pennebakerlinguistic}
    Pennebaker, J.~W., Booth, R.~J., Boyd, R.~L., and Francis, M.~E.
    \newblock (2015).
    \newblock Linguistic inquiry and word count: Liwc2015.
    
    \bibitem[\protect\citename{Redlawsk}2002]{Redlawsk_hotcognition}
    Redlawsk, D.~P.
    \newblock (2002).
    \newblock Hot cognition or cool consideration? testing the effects of motivated
      reasoning on political decision making.
    \newblock {\em The Journal of Politics}.
    
    \bibitem[\protect\citename{Shilton and Sayles}2016]{shilton2016we}
    Shilton, K. and Sayles, S.
    \newblock (2016).
    \newblock " we aren't all going to be on the same page about ethics": Ethical
      practices and challenges in research on digital and social media.
    \newblock In {\em 2016 49th Hawaii International Conference on System Sciences
      (HICSS)}, pages 1909--1918. IEEE.
    
    \bibitem[\protect\citename{Sirrianni \bgroup et al.\egroup
      }2020]{sirrianni-etal-2020-agreement}
    Sirrianni, J., Liu, X., and Adams, D.
    \newblock (2020).
    \newblock Agreement prediction of arguments in cyber argumentation for
      detecting stance polarity and intensity.
    \newblock In {\em Proceedings of the 58th Annual Meeting of the Association for
      Computational Linguistics}, pages 5746--5758, Online, July. Association for
      Computational Linguistics.
    
    \bibitem[\protect\citename{Sobkowicz}2012]{Sobkowicz2012DiscreteMO}
    Sobkowicz, P.
    \newblock (2012).
    \newblock Discrete model of opinion changes using knowledge and emotions as
      control variables.
    \newblock {\em PLoS ONE}, 7.
    
    \bibitem[\protect\citename{Somasundaran and
      Wiebe}2010]{somasundaran-wiebe-2010-recognizing}
    Somasundaran, S. and Wiebe, J.
    \newblock (2010).
    \newblock Recognizing stances in ideological on-line debates.
    \newblock In {\em Proceedings of the {NAACL} {HLT} 2010 Workshop on
      Computational Approaches to Analysis and Generation of Emotion in Text},
      pages 116--124, Los Angeles, CA, June. Association for Computational
      Linguistics.
    
    \bibitem[\protect\citename{Stab \bgroup et al.\egroup
      }2018]{stab-etal-2018-cross}
    Stab, C., Miller, T., Schiller, B., Rai, P., and Gurevych, I.
    \newblock (2018).
    \newblock Cross-topic argument mining from heterogeneous sources.
    \newblock In {\em Proceedings of the 2018 Conference on Empirical Methods in
      Natural Language Processing}, pages 3664--3674, Brussels, Belgium,
      October-November. Association for Computational Linguistics.
    
    \bibitem[\protect\citename{Strandberg \bgroup et al.\egroup
      }2018]{fd84e41afabf4ffdbe90fb4f3a398eaf}
    Strandberg, K., Himmelroos, S., and Gr{\"o}nlund, K.
    \newblock (2018).
    \newblock Do discussions in like-minded groups necessarily lead to more extreme
      opinions?: Deliberative democracy and group polarization.
    \newblock {\em International Political Science Review}.
    
    \bibitem[\protect\citename{Taber and
      Lodge}2006]{https://doi.org/10.1111/j.1540-5907.2006.00214.x}
    Taber, C.~S. and Lodge, M.
    \newblock (2006).
    \newblock Motivated skepticism in the evaluation of political beliefs.
    \newblock {\em American Journal of Political Science}, 50(3):755--769.
    
    \bibitem[\protect\citename{Thomas \bgroup et al.\egroup
      }2006]{thomas-etal-2006-get}
    Thomas, M., Pang, B., and Lee, L.
    \newblock (2006).
    \newblock Get out the vote: Determining support or opposition from
      congressional floor-debate transcripts.
    \newblock In {\em Proceedings of the 2006 Conference on Empirical Methods in
      Natural Language Processing}, pages 327--335, Sydney, Australia, July.
      Association for Computational Linguistics.
    
    \bibitem[\protect\citename{Townsend and Wallace}2016]{townsend2016social}
    Townsend, L. and Wallace, C.
    \newblock (2016).
    \newblock Social media research: A guide to ethics.
    \newblock {\em University of Aberdeen}, 1:16.
    
    \bibitem[\protect\citename{Vicario \bgroup et al.\egroup
      }2016]{delvicario2016modeling}
    Vicario, M.~D., Scala, A., Caldarelli, G., Stanley, H.~E., and Quattrociocchi,
      W.
    \newblock (2016).
    \newblock Modeling confirmation bias and polarization.
    
    \bibitem[\protect\citename{Walker \bgroup et al.\egroup
      }2012a]{walker-etal-2012-corpus}
    Walker, M., Tree, J.~F., Anand, P., Abbott, R., and King, J.
    \newblock (2012a).
    \newblock A corpus for research on deliberation and debate.
    \newblock In {\em Proceedings of the Eighth International Conference on
      Language Resources and Evaluation ({LREC}-2012)}, pages 812--817, Istanbul,
      Turkey, May. European Languages Resources Association (ELRA).
    
    \bibitem[\protect\citename{Walker \bgroup et al.\egroup
      }2012b]{Walker_thatsyour}
    Walker, M.~A., Anand, P., Abbott, R., Tree, J. E.~F., Martell, C.~H., and King,
      J.
    \newblock (2012b).
    \newblock That is your evidence?: Classifying stance in online political
      debate.
    
    \bibitem[\protect\citename{Williams \bgroup et al.\egroup
      }2017]{williams2017towards}
    Williams, M.~L., Burnap, P., and Sloan, L.
    \newblock (2017).
    \newblock Towards an ethical framework for publishing twitter data in social
      research: Taking into account users’ views, online context and algorithmic
      estimation.
    \newblock {\em Sociology}, 51(6):1149--1168.
    
    \bibitem[\protect\citename{Zhu \bgroup et al.\egroup }2019]{wrap116715}
    Zhu, L., He, Y., and Zhou, D.
    \newblock (2019).
    \newblock Neural opinion dynamics model for the prediction of user-level stance
      dynamics.
    \newblock {\em Information Processing and Management}, March.
    
    \end{thebibliography}
\end{document}